\documentclass[letterpaper, 10 pt, conference]{ieeeconf}  

\IEEEoverridecommandlockouts                              

\RequirePackage[2014/01/01]{latexrelease}
\usepackage{graphics} 
\usepackage{epsfig} 
\usepackage{mathptmx} 
\usepackage{times} 
\usepackage{amsmath} 
\usepackage{amssymb}  
\usepackage{array}

\title{\LARGE \bf
Unconstrained Road Marking Recognition with Generative Adversarial Networks
}

\author{
{Younkwan Lee, Juhyun Lee, Yoojin Hong, YeongMin Ko, Moongu Jeon} \\
{Machine Learning and Vision Laboratory, GIST, Gwangju, South Korea} \\
{\tt\small \{brightyoun, leejuhyun, yoojinhong, koyeongmin, mgjeon\}@gist.ac.kr}
}

\begin{document}

\maketitle
\thispagestyle{empty}
\pagestyle{empty}

\begin{abstract}

Recent road marking recognition has achieved great success in the past few years along with the rapid development of deep learning. Although considerable advances have been made, they are often over-dependent on unrepresentative datasets and constrained conditions. In this paper, to overcome these drawbacks, we propose an alternative method that achieves higher accuracy and generates high-quality samples as data augmentation. With the following two major contributions: 1) The proposed deblurring network can successfully recover a clean road marking from a blurred one by adopting generative adversarial networks (GAN). 2) The proposed data augmentation method, based on mutual information, can preserve and learn semantic context from the given dataset. We construct and train a class-conditional GAN to increase the size of training set, which makes it suitable to recognize target. The experimental results have shown that our proposed framework generates deblurred clean samples from  blurry ones, and outperforms other methods even with unconstrained road marking datasets.

\end{abstract}

\section{INTRODUCTION}

    Road marking recognition is one of fundamental and crucial tasks for self-driving cars. From the perception system in a self-driving car, the ability to automatically recognize road markings on the road surface is one of the key steps toward understanding road conditions. Therefore, road marking recognition is indispensable to ensure a safe and reliable navigation of self driving cars through the road way.  
    
    \begin{figure}[thpb]
      \centering
        \includegraphics[width=3.2in]{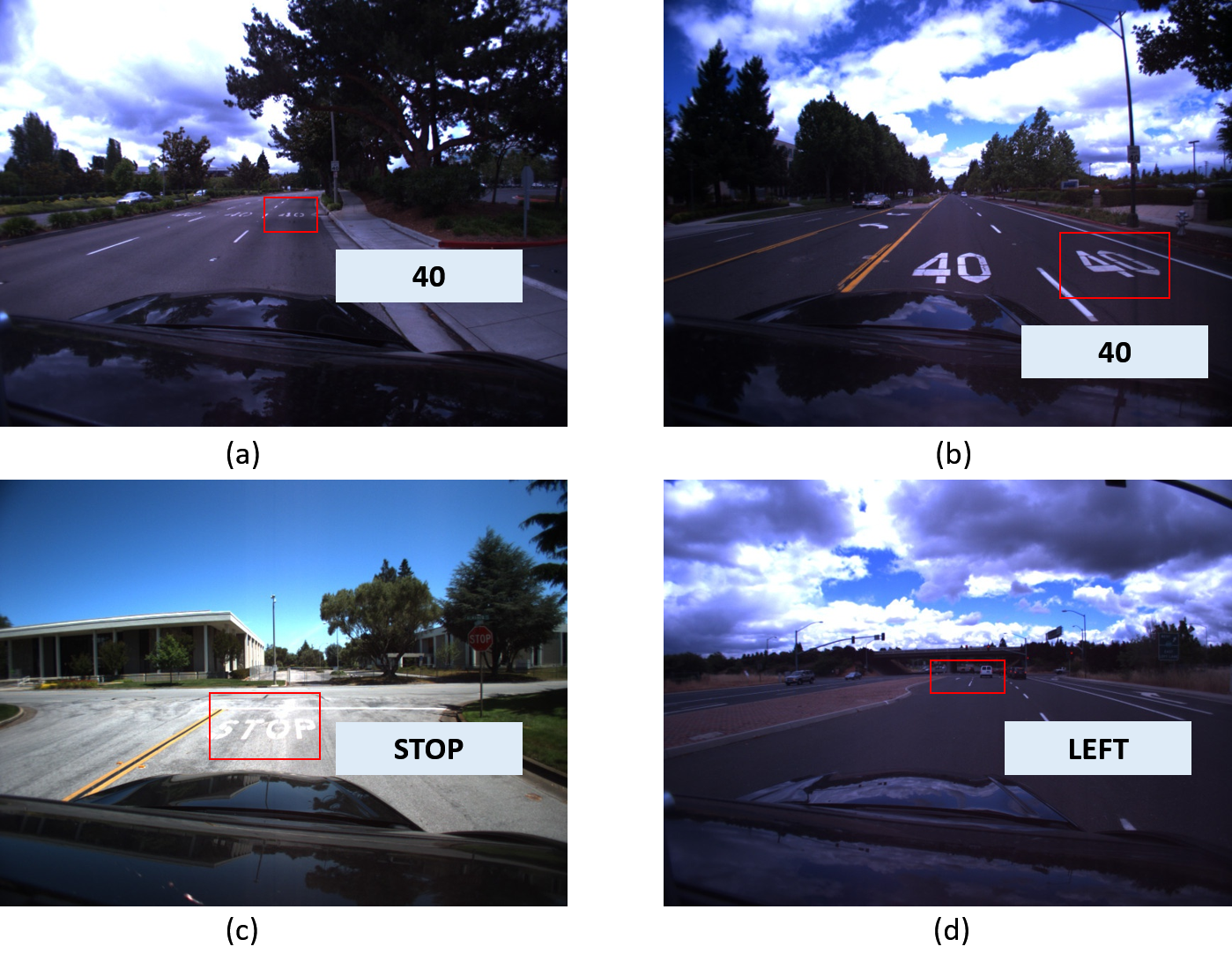}
      \caption{Algorithm testing results during common challenging conditions: (a) low-resolution and blurred image, (b) distortion by angle, (c) intense illumination, (d) long distance from camera and blurred image. Blue rectangular box is class of red bounding box.}
      \label{figurelabel}
    \end{figure}

    Over the past decade, there has been considerable research in the area of road marking recognition. Currently, huge progress of road marking recognition has been made with emergence of deep convolutional neural networks. However, it is still a challenging problem on unconstrained environments such as uneven illumination, distortion, low resolution, and extreme weather conditions. In addition, existing road marking datasets are limited in diversity and quantity. Because most existing methods \cite{c1, c2} often evaluate their approaches with unrepresentative datasets collected on extremely limited road surface, they might work well only under some controlled conditions. However, the above-mentioned unconstrained road environments are very common in real world, as shown in Fig 1. For those reasons, existing methods are fairly unsuitable for real-world scenarios due to inconsistent performance.

    To deal with such difficulties, we propose an end-to-end convolutional neural network framework. It is similar to generative adversarial networks (GAN) \cite{c3} framework, and capable of deblurring, augmentation, and classification of road markings with low error rates based on the adversarial training strategy. There are two sub-networks in our model: a generator network and a discriminator network. In the generator network, our focus is on both deblurring the positive samples and distorting the negative samples simultaneously from the unconstrained inputs. Most of existing works are tailored to recover under the assumption that the input is always positive sample. As a result, they tend to mislead the output to be similar to the positive sample when the input is even a negative sample. Because of the highly unconstrained scene of road marking detection, such an approach that can lead to false positive errors is very far from an unconstrained reality. Therefore, we propose a deblurring network to restore some blurry regions for positive samples only and decimate the entire region for negative samples. The discriminator network is composed of two branches: one discriminates real road marking from fake one, and the other categorizes various road markings for the task of object detection. Since scarcity of dataset causes overfitting, we propose a data augmentation module which is a generator network based on mutual information, to capture a much larger invariance space and preserve semantic context. Training the classifier with the augmented data would improve classification tasks significantly. Benefiting from the augmentation module, in the testing phase, the discriminator network performs as the actual road marking classifier, while the generator network improves the performance of the classifier by adequately deblurring the positive samples and decimating any given negative (not road marking) samples.

 \begin{figure*}[thpb]
\begin{center}
   \includegraphics[width=7in]{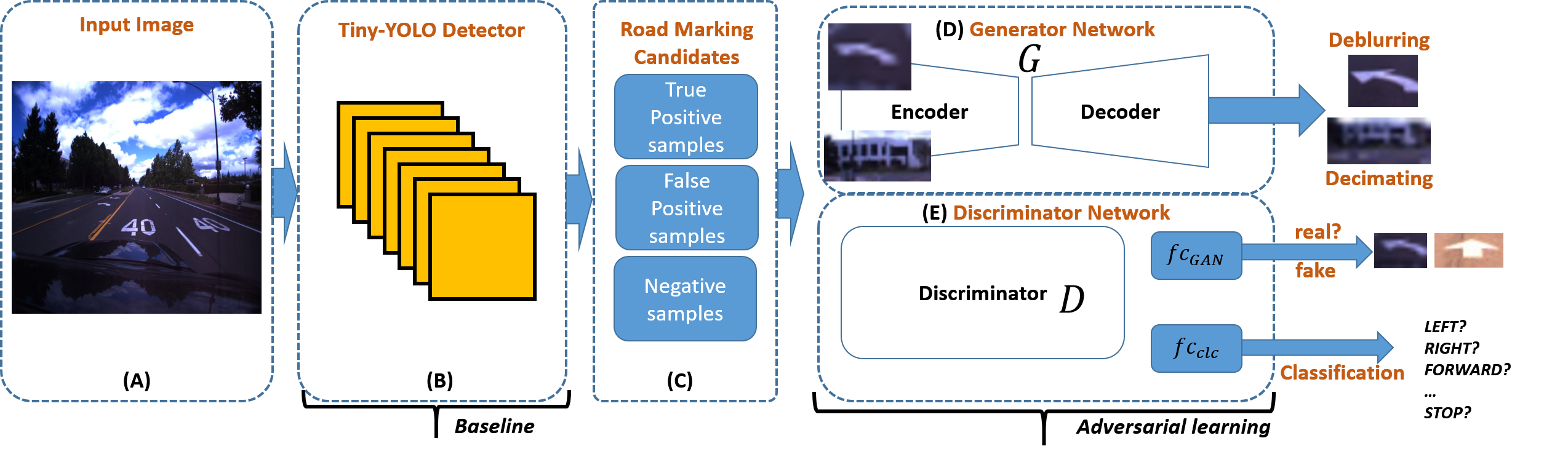}
\end{center}
   \caption{The pipeline of the proposed road marking recognition system. (A) The images are fed into the network; (B) Tiny-YOLO detector is our baseline \cite{c4}; (C) The outputs from the baseline are composed of the positive data and negative data; (D, E) The adversarial framework is jointly trained by two branch loss.}
\label{fig:asdasd}
\end{figure*}

    Our main contributions are summarized below: First, we propose a novel end-to-end deep network to generate the clean images from blurred samples using adversarial training for road marking recognition. Second, we apply a data augmentation module to producing samples which are different to the original samples in the dataset, but enhance sample quality for generalization performance. Finally, performance of our proposed method in distinguishing a road marking from an unconstrained conditions, is best among compared methods in the most experiments, especially on the unconstrained dataset.

\section{RELATED WORKS}
Existing methods in the literature for road marking recognition can roughly be categorized into two approaches, namely traditional feature based approaches and deep feature based approaches.

\textbf{Traditional feature based approaches.} Hand-crafted features (such as HOG) are widely used in this field, were introduced for road marking recognition in \cite{c1,c5,c6,c7}. Typically, Tao \textit{et al.} \cite{c1} noticed that the problem of road marking detection and classification has to be separated undoubtedly. Therefore, they introduced Inverse Perspective Mapping(IPM) method \cite{c6} to prevent the perspective effect. Also, FAST feature \cite{c84} is utilized to capture interests of feature point using HOG features \cite{c9} at three different scales. Then, Extremal Regions (ER) are employed to generate multiple road marking ROIs, which are then improved by Maximally Stable Extremal Regions (MSERs) \cite{c10}. Instead of high-computation cost feature, Li \textit{et al.} \cite{c6} extract ROIs from IPM images and operate low-cost feature extraction computation to recognize target road markings. However, this work not only abandons quantitative results but also has a drawback in that new road marking classes must be newly learned. Greenhalgh \textit{et al.} \cite{c7} utilized a Support Vector Machine (SVM) by extracting HOG features. However, this work only recognizes text-based road markings and cannot be applied to iconic road marking.

\textbf{Deep feature based approaches.} More recently, neural networks have been successfully introduced for road marking recognition. Kheyrollahi \textit{et al.} \cite{c11} proposed an approach to this problem based on the extraction of robust road marking features via a novel pipeline of inverse perspective mapping and multi-level binarisation. The approach is shown to operate successfully over a range of unconstrained environments such as lighting, weather and road surface conditions, applying fully connected neural networks. With the rise of deep neural networks and their success \cite{c12, c13} in computer vision applications, Ahmad \textit{et al.} \cite{c2} trained five different CNN architectures with variable number of convolution/max-pooling and fully connected layers. Different from public dataset of general objects, road marking public benchmark has a small amount of examples less than 10,000 samples. To resolve this issue, they introduced data augmentation to enhance the size of dataset using in-plane rotation of the original images. However, standard linear data augmentation produces only limited plausible alternative data.

We have used the convolutional neural network and deep learning for feature extraction \cite{c11}, \cite{c2}. Furthermore, our work is also related to data augmentation. \cite{c2} first utilized a data augmentation method of road marking. Different from the above method, our framework designs and trains a generative model to do data augmentation, it can be applied to even small and unconstrained data.

\section{PROPOSED METHOD}
    In this section, we describe our approach to road marking recognition. First, we illustrate how the vanilla GAN \cite{c14} and InfoGAN \cite{c15} can be applied to this problem. Next, we present the proposed whole architecture of our method, as shown in Fig 2. Lastly, we introduce each part of our network in details and represent the loss functions for training the generator network and discriminator network respectively, and describe how our approach improves the performance of road marking recognition.

\subsection{GAN} GAN framework is composed of two networks: a generator network $G$, and a discriminator network $D$. The generator network $G$ and discriminator network $D$ are trained simultaneously by playing a two-player minimax game. Thus, the generator $G$ attempts to minimize the differences between the real samples and the fake samples generated by $G$ to fool the discriminator $D$. On the other hand, the discriminator $D$ aims at maximizing such differences for distinguishing real and fake. The GAN objective function can be formulated as the following minimax objective function:
        \begin{equation}
        \begin{split}
            \min_{\theta_G} \max_{\theta_D} V(D,G) = &\mathbb{E}_{x\sim p_{t}~(x)}[log D(x)] \\
            & + \mathbb{E}_{z\sim p_{z}(z)}[log(1-D(G(z)))]
        \end{split}
        \end{equation}
    where $p_{t}$ is the real data distribution observation from $x$ and $p_{z}$ is the fake data distribution observation from a random distribution $z$. GAN alternatively optimizes two networks, which compete each other. Therefore, the conclusion to play the minimax game can be that the probability distribution ($p_{z}$) generated by the generator $G$ exactly matches the data distribution ($p_{t}$). After all, the discriminator $D$ will not be able to distinguish between sampling distribution from the generator $G$ and real data distribution. At this time, for the fixed generator, the optimal discriminator function is as follows:
        \begin{equation}
            D_{G}^*(x) = \frac{p_{t}(x)}{p_{t}(x) + p_{z}(x)}.
        \end{equation}
    
    More recently, associated to mutual information maximization, we further design GAN framework to cope with the problem of trivial codes. By maintaining information relating to the semantic features of the data, the network is able to learn more meaningful hidden representations for deblurring and classification. The mutual information can be formulated as follows:
    \begin{equation}
    {I(c;G(z,c))}=H(c) - H(c|G(z,c)),
    \end{equation}
    where $c$ denotes the salient structured semantic features of the data distribution by $c_1, c_2, ..., c_N$, $H(\cdot)$ and $H(\cdot|\cdot)$ represent the marginal and conditional entropy respectively, and $I(\cdotp;\cdot)$ denotes the mutual information. we maximize the mutual information because the information in the latent code $c$ should not be lost in the generation process. Therefore, the minimax game of a joint network ${G}$+${D}$ is formulated as follows:
        \begin{equation}
        \begin{split}
            & \arg \min_{\theta_G} \max_{\theta_D} V(D,G) = \mathbb{E}_{(X,y)\sim p_{t}}[log D(X, y)] +\\
            & \mathbb{E}_{(\tilde{X},y)\sim p_{t}+\mathcal{N}_\sigma}[log(1-D(G(\tilde{X}, y)))] - \lambda{I(y;G(\tilde{X}, y))},
        \end{split}
        \end{equation}
    where $\mathcal{N}$ is the noise model from the normal distribution with standard deviation $\sigma$, $X$ denotes road marking candidates with clean image, $\tilde{X}$ represents the road marking candidates with blurred noise, and $y$ is the road marking label. In the discriminator network $D$, we distinguish the generated $vs.$ true images and road marking classification jointly.

\begin{figure}[t]
\begin{center}
   \includegraphics[width=1\linewidth]{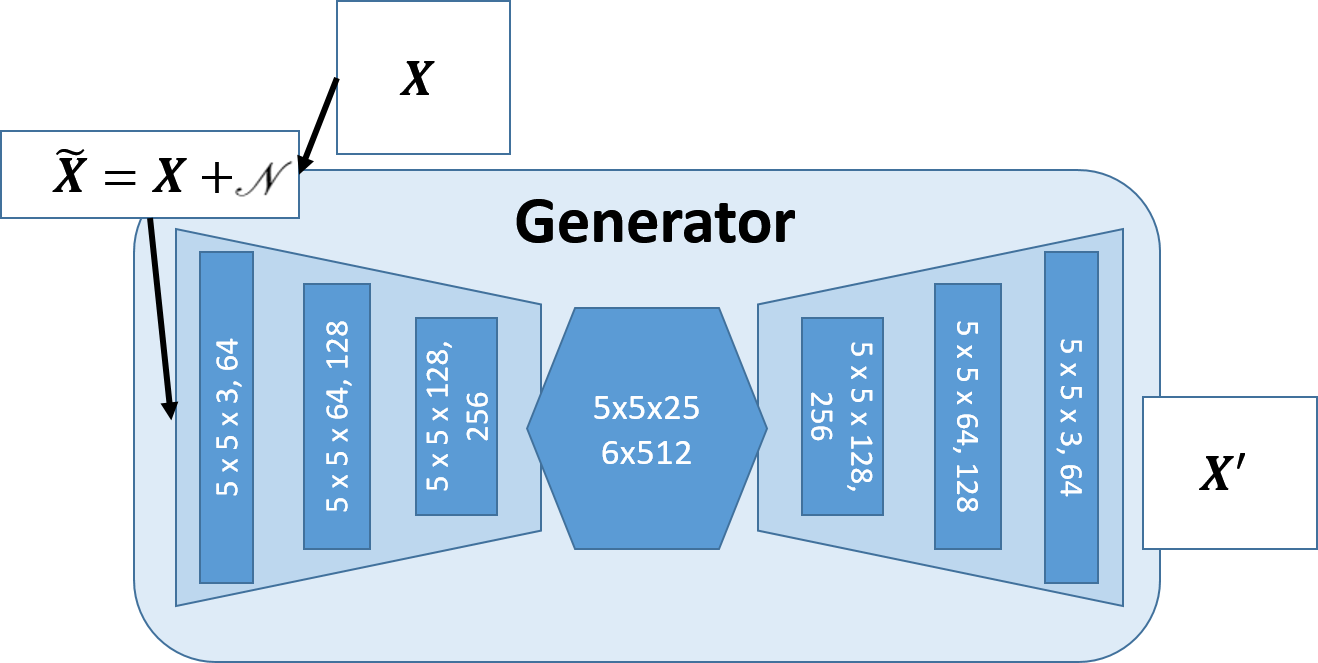}
\end{center}
   \caption{${G}$ network architecture, composed of encoding (first part) and decoding (second part) layers.}
\label{fig:R}
\end{figure}

     \begin{figure}[b]
\begin{center}
   \includegraphics[width=0.8\linewidth]{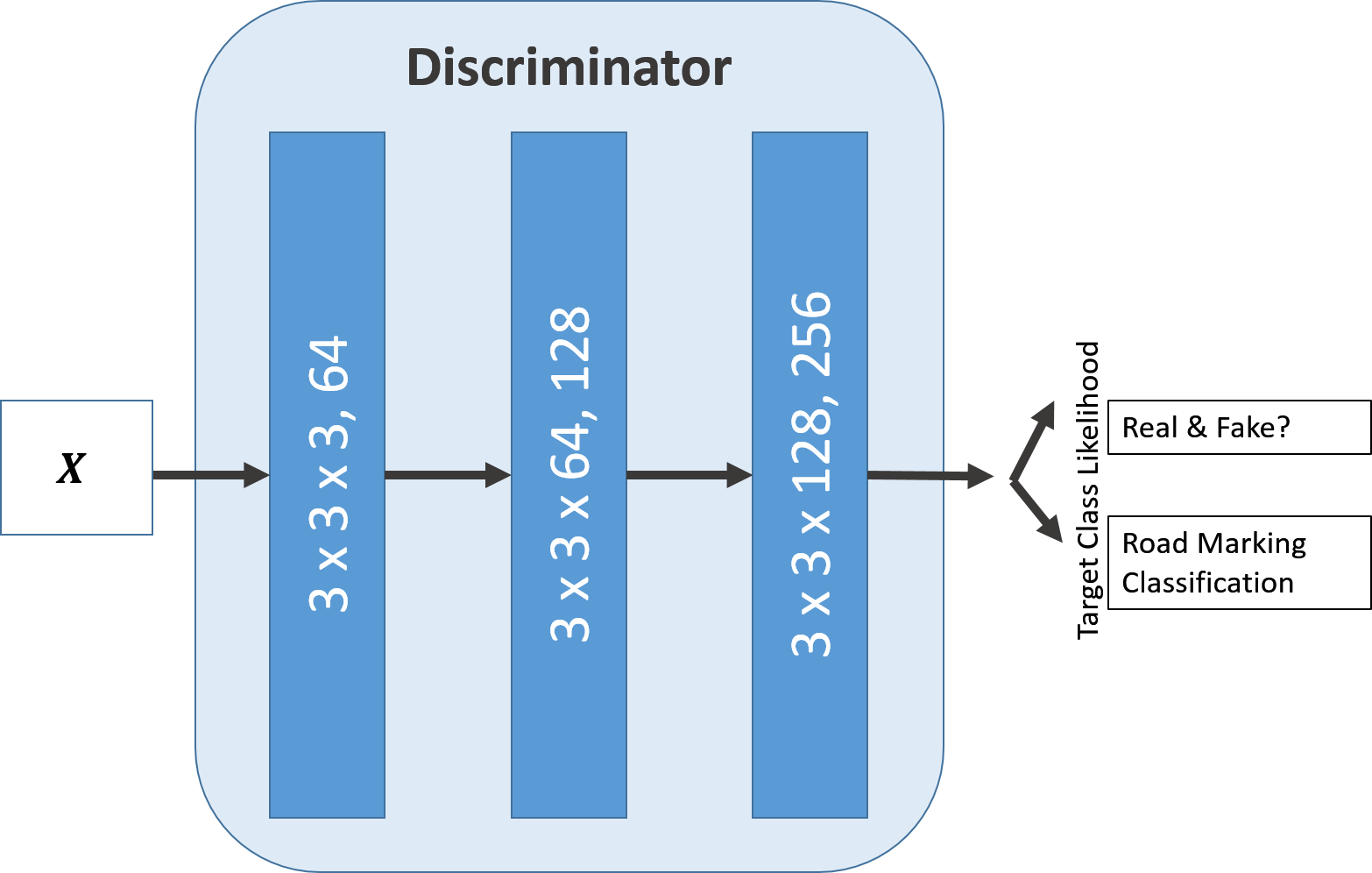}
\end{center}
   \caption{${D}$ network is composed of three convolutional layers with two parallel branches.}
\label{fig:D}
\end{figure}

\subsection{Proposed Network Architecture} Our generative adversarial networks framework includes two sub-networks: (1)the generator network and (2)the discriminator network. Below are the details of our network architectures.

    \textbf{Generator network} Inspired by the recent success of image reconstruction \cite{c16}, we implement the generator model $G$ using a fully convolutional auto-encoder, as shown in Fig 3. Since the input image sampled from the unconstrained scenes is very corrupted, converting into the clean one is quite useful for road marking classification. Along with this idea, our $G$ network trains deblurring mappings from corrupted positive images (\textit{i.e.} road marking) to the clean ones, while it learns decimating mappings from the outliers (\textit{i.e.} not road marking), making it easier for the classification to distinguish the outliers from a unconstrined samples in an end-to-end fashion. As a result, $G$ can eliminate noise only in positive samples while preserving the abundant image details. Eventually, $G$ learns the target representation to deliver an artifact-free and detail-preserving property to the positive samples and corruption to the negative samples.
    
    \textbf{Discriminator network} We employ a sequence of convolution layers as our discriminator network $D$, as shown in Fig 4. In existing GAN frameworks, the discriminator network has only one branch that distinguishes the real and generated samples. Instead, in our discriminator model, $D$ has two parallel branches, namely, a classifier of real/fake sample $fc_{GAN}$ and a road marking classifier $fc_{clc}$. Fig.4 shows the details of this network's architecture. The output of $fc_{GAN}$ branch is the probability of the input being a real, and the output of $fc_{clc}$ is the softmax probability of the input being categories of the road markings.
    
\subsection{Overall Loss Function} 

    To train the model, we calculate the loss function ${L}_{{G}+{D}}$ of the joint network ${G}$+${D}$ using only the positive samples. When we optimize our generator network, we adopt the pixel-wise loss instead of feature matching loss, defined as:
    \begin{equation}
    {L}_{MSE}=\| X-X'\|^{2}.
    \end{equation}
    where $X'$ is the output of $G$. However, the MSE optimization problem often leads to a lack of the sharpening effect, which results in over-smoothing effect that still causes blurring. 
    
    To resolve this issue, we also introduce the classification loss, which promotes $G$ to make as large as possible the sharpening effect. The formulation of classification loss is
    	\begin{equation}
        	L_{clc} = { (\log(y_i - D(G(\tilde{X}))) + \\
        	\log(y_i - D(X)))},
    	\end{equation}
    where ${y}$ indicates the corresponding labels of the image and $i$ represents the number of classes. Consequently, the classification loss provides an enhancement in the recovered images than when not in use.
    
    Based on above functions, the model is optimized to minimize the loss function:
    \begin{equation}
    {L}={L}_{{G}+{D}} + \lambda {L}_{MSE} + {L}_{clc},
    \label{eq:sum_loss}
    \end{equation}
    where $\lambda > 0$ is a trade-off weights that controls the relative importance of the MSE loss. For finding a proper hyper-parameter $\lambda$ of training $L$, we have conducted a number of experiments and finally introduce an appropriate value 0.05. Based on this finding, we have stopped the training of networks, when ${G}$ can  reconstruct its input with minimum error ($\|X-X'\|^{2} < \rho$, where $\rho$ is a small positive number).
    
    After the training of networks, we analyze the behavior of each input. For any given positive sample that follow $p_t$, $G$ is trained and operates denoising task, and its output will be a clean version of the input data. On the other hand, for any given outlier sample that does not follow $p_t$, $G$ is disturbed and cannot reconstruct it correctly. Therefore, $G$ plays a role of denoising as well as decimating for non-critical negative samples and thereby causing negative samples away from the possibility of false positive error.
    
 \begin{figure}[b]
\begin{center}
   \includegraphics[width=1\linewidth]{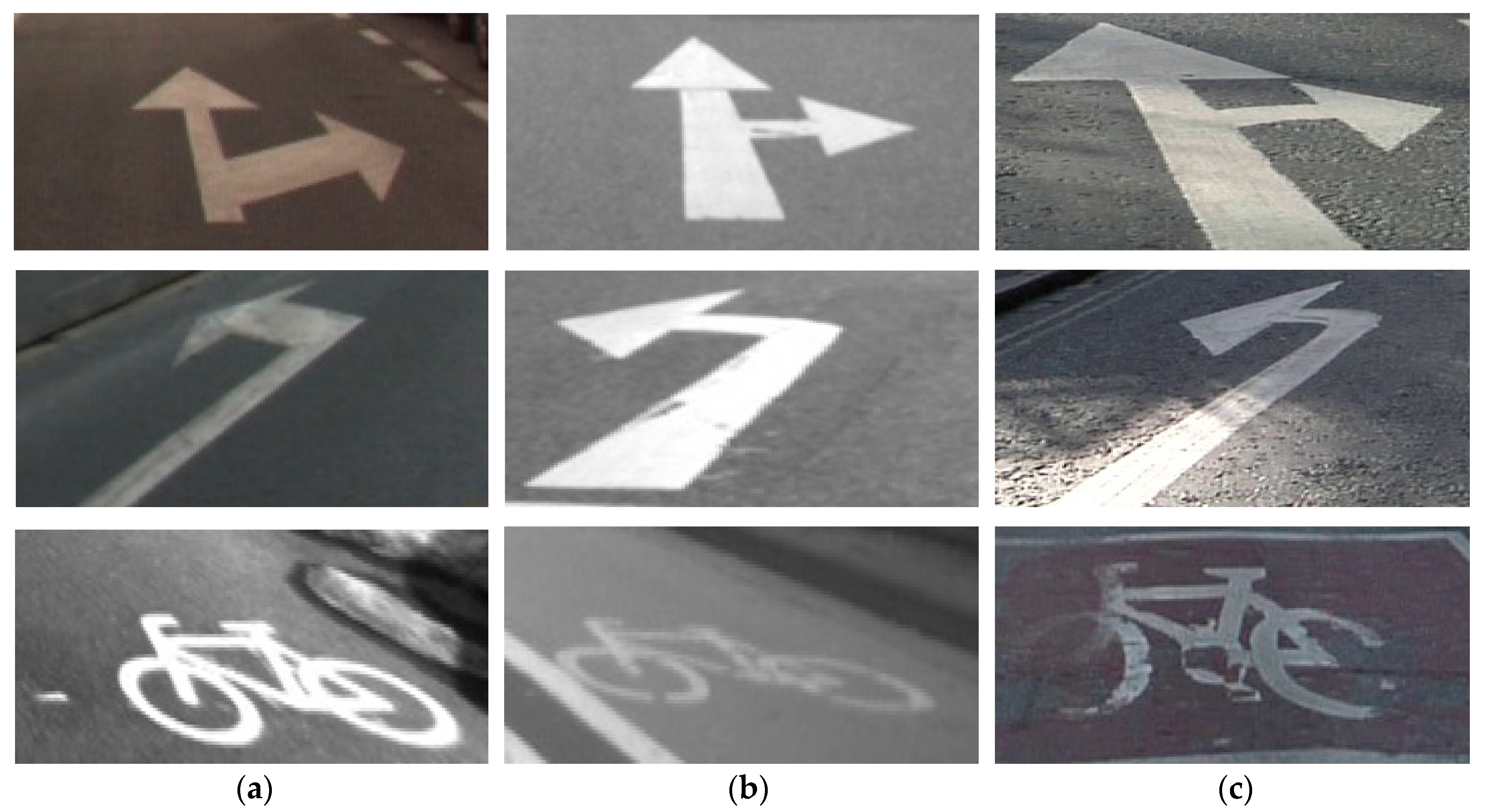}
\end{center}
   \caption{Results of data augmentation using mutual information based GAN. There is no real image in this figure.}
\label{fig:D}
\end{figure}

\begin{table}[]
\caption{Number of instances for each class of road markings.}
\begin{tabular}{c|cc}
\hline
\textbf{Road Marking Classes} & \textbf{Orininal instances} & \textbf{Augmented instances} \\ \hline
35                            & 112                         & 300                          \\
40                            & 69                          & 300                          \\
FORWARD                       & 86                          & 300                          \\
LEFT                          & 705                         & 300                          \\
PED                           & 54                          & 300                          \\
RAIL                          & 90                          & 300                          \\
RIGHT                         & 101                         & 300                          \\
STOP                          & 49                          & 300                          \\
XING                          & 64                          & 300                          \\
NULL                          & 2,700                       & 2,000                          \\ \hline
\textbf{Total}                & \textbf{4,030}              & \textbf{4,700}              
\end{tabular}
\end{table}

\begin{table}[]
\caption{The classification results according to data augmentation models. For fair comparison, we use augmented dataset as trainingset.}
\begin{tabular}{c|c}
\hline
\textbf{Augmentation Model} & \textbf{Classification Performance} \\ \hline
Ours w/o augmentation            & 99.2\%                              \\
Ours + augmentation (Vanilla GAN)                         & 93.2\%                              \\
Ours + augmentation (InfoGAN)                     & 98.8\%                             
\end{tabular}
\end{table}

\begin{figure}[b]
\begin{center}
   \includegraphics[width=1\linewidth]{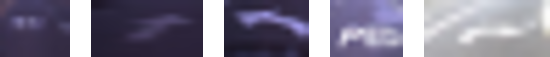}
\end{center}
   \caption{Some examples of the blurry road marking generated by the classical GAN.}
\label{fig:001}
\end{figure}

 \begin{figure}[b]
\begin{center}
   \includegraphics[width=1\linewidth]{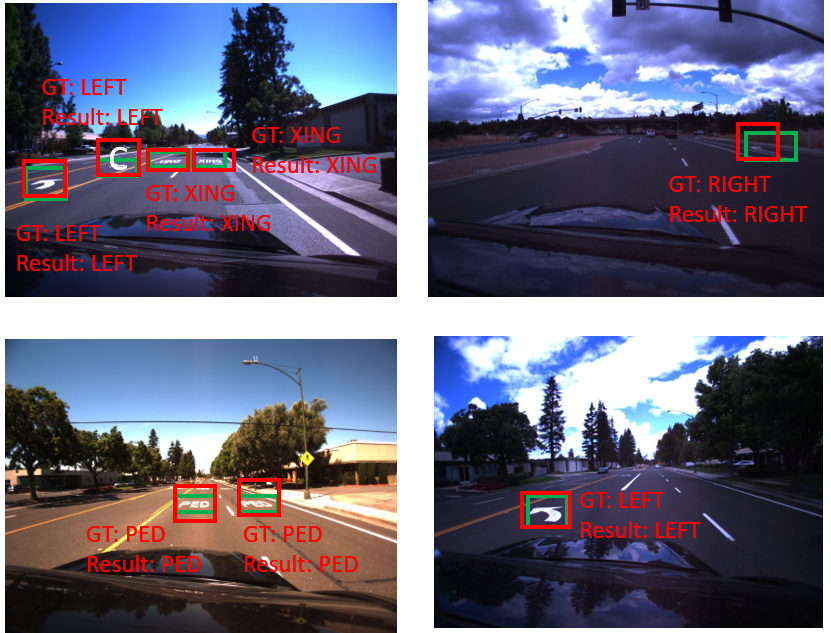}
\end{center}
   \caption{Qualitative recognition results of our proposed method. Green bounding boxes are ground truth annotations of road marking and red bounding boxes are the results from our method.}
\label{fig:D}
\end{figure}

\section{GAN-BASED DATA AUGMENTATION}
The data augmentation model can be used in the latent code of $G$ network based on mutual information. Consider a source domain consisting of data $D=\{X_1,X_2,\ldots,X_{N_D}\}$ and corresponding target classes $\{y_1,y_2,\ldots,y_{N_D}\}$. Given a data point ${X, y}$ we could obtain a meaningful representation of the data point, which encapsulates the information needed to generate other related data. As a result, the latent code of $G$ learns and preserves the class-specific semantic context, as shown in Fig 5.

To illustrate the advantage of InfoGAN based augmentation, we also use the classical GAN. For the classical GAN, we train a fully convolutional generator without using any labeled data. While the classical GAN optimizes its loss function, its generator visually generates almost spurious data, as shown in Fig 6. From Table II, the classification accuracy of road markings by GAN makes it inferior to that when not done. This is because of the scarcity and low quality of the dataset, and most importantly, it does not learn the semantic representation of class-specific samples at all. In other words, maximizing the mutual information of a class-specific image and its dependent latent code contributes to consistent and high-quality data generation. Further analysis on the influence of data augmentation is presented in Table II.

    \setlength{\tabcolsep}{10pt}
    \begin{table*}
    \begin{center}
    \caption{Comparison of results of our method with other state-of-the-art methods.}
    \label{table:headings}
    \begin{tabular}{l|cc}
    \hline
    Method $\qquad\qquad\qquad$& Classifier Accuracy  & \multicolumn{1}{c|}{Detection and Recognition Accuracy}\\
    \hline\hline
    \noalign{\smallskip}
    Shallow CNN \cite{c12} & 94.7\% & 90.8\% \\ 
    PCANet+SVM \cite{c17} & 99.1\% & 93.1\%\\  
    PCANet+Logistic Regression \textit{et al.} \cite{c17} & 98.9\% & 92.4\%\\ \hline\hline 
    Baseline (YOLO v2) \cite{c4} & 94.6\% & 89.8\%\\ 
    Ours without deblurring network & 95.7\% & 90.8\%\\ 
    Ours without classification loss & 97.9\% & 93.4\%\\ 
    Ours without augmentation dataset & 99.2\% & 93.6\%\\ 
    \textbf{Ours} & \textbf{98.8\%} & \textbf{95.3\%}\\ 
    \hline
    \end{tabular}
    \end{center}
    \end{table*}

\section{EXPERIMENTAL RESULTS}

\subsection{Setup}
All the reported implementations are based on TensorFlow framework, and our method has been done on the NVIDIA TITAN X GPU. First of all, we use the YOLO-v2 \cite{c4} through pre-trained model on COCO \cite{c18} as our baseline so that we trained road marking samples by fine-tuning their network parameters. 

Also, to avoid the premature convergence of the discriminator network, the generator network is updated more frequently than original one. In addition, higher learning rate is applied to the training of the generator. For stable training, we use Adam optimizer \cite{c19} with a high momentum term. The weights of the discriminator network in all parallel fully connected layers are initialized from the standard Gaussian distribution with zero-mean, a standard deviation of $0.01$ and the constant $0$ as the bias in all layers. All models are trained on loss function for first 10 epochs with initial learning rate of $10^{-4}$. After that, we set the learning rate to a further reduced $10^{-5}$ for the remaining epochs. Finally, batch normalization \cite{c20} is used in all layers of generator and discriminator, except the last layer of the $G$ and the first layer of the $D$. 

\subsection{Dataset}
\textbf{Original Dataset} 
We use the road marking dataset provided by Tao et al. \cite{c1} for all experiments. As stated in \cite{c1}, there are 1,330 road images of different classes in the dataset. For the object detection task, we randomly extract 2,700 background images that not belong to any class, namely the "NULL" class. The distribution of types in the dataset is shown in Table I. The data include information about the pixel coordinates for each road marking. We divide benchmark dataset into 2,418 training samples and 1,612 test samples (60\% : 40\%). We will randomly rearrange the images so that the training and test images are randomly selected without overlapping. 

\textbf{Augmentation Dataset} We adopt training samples of each class from the original dataset for augmentation model. Our augmentation model outputs 2,700 samples of the full 9 classes as the training set. Moreover, we also generate outlier samples from "NULL" class in the original dataset which is randomly extracted, resulting in 2,000 images. Therefore, this dataset contains 10 classes with a total of 4,700 images. 

\subsection{Comparison with Other Methods}
We compare our proposed method with state-of-the-art methods on \cite{c1} public road marking benchmark. We also measure the performance of our classification branch in the discriminator by 7,118 images for training and 1,612 images for test. From Table III, we see that the recognition performance drops by about 4\% without the deblurring network. The reason is that the reconstructed images derived by MSE loss and classification loss are over noisy. Since a detailed representation is not learned, it causes confusion and mistakes in the discriminator network. Furthermore, we see that the recognition performance increases by about 1\% with the classification loss. As discussed, the classification loss enhances the generator to recover meaningful information for accurate classification.

Interesting to note, even though no augmentation method shows better classifier accuracy upon training compared to other methods, our method including data augmentation performs better during the detection and recognition test. From detection and recognition accuracy of Table III, it reassures the claim that the probability to belong to a specific class that is derived from low confidence score, is not necessarily confident. Instead, using data augmentation based on InfoGAN with the proposed adversarial network ensures that only the most appropriate regions remain.

From Table III, our method demonstrates the highest detection and recognition results, outperforming the state-of-the-art road marking method by more than 2\% in recognition accuracy. Based on this observation, the advancement of our performance mainly comes from three contributions: (1) our generator network $G$ learns a deblurred image for positive samples, while it distorts and decimates image that do not contain the road marking information. Consequently, this affects the discriminator $D$ better to distinguish the testing samples; (2) the classification loss $L_{clc}$ boosts the generator $G$ to capture a meaningful semantic representation for easier classification; (3) data augmentation based on mutual information enables effective our model training even in low-quality target domains. As the data augmentation module does not depend on the classes themselves, it learns the cross-class semantic feature. As a result, existing methods fail to recognize unconstrained road markings. However, our method is most notable, which clearly demonstrates the effectiveness of our method on unconstrained road marking recognition.

Visual results of our road marking detection and recognition are undoubtedly reassuring and can be observed in Fig 7. Currently, our algorithm is sensitive to occlusion by vehicles on the road, since the information is not enough to provide consistent classification score. Additionally, the algorithm fails when the road markings are merged with the lane segment, due to human error during the road marking painting process. This happens because the baseline detector extracts true road markings together with a lane segment into one region since they appear to be connected.

\section{CONCLUSIONS}
In this paper, we presented a robust GAN framework for road marking detection and classification. Specifically, our framework is proven to be robust to various unconstrained environments. Unlike existing methods, we design a generator network to directly restore a deblurred image from a blurry positive one, while distort a decimated image from a negative one. In addition, we have proposed a two classification branch to the discriminator network, which can distinguish the fake/real and road marking categories including negative class, resulting in accuracy improvement and convenient evolvement. Additionally, by applying data augmentation based on mutual information it leads to better performance than other state-of-the-art augmentation methods. We have demonstrated its general applicability by incorporating modules with Tiny-YOLO as baseline. Finally, our approach can be integrated easily to ADAS and ADS systems and, extended to instance segmentation for exact localization of road markings in the future.

\addtolength{\textheight}{-12cm}   




\section*{ACKNOWLEDGMENT}

This work was partly supported by the ICT R\&D program of MSIP/IITP. (2014-0-00077, Development of global multi target tracking and event prediction techniques based on real-time large-scale video analysis) and GIST Autonomous Vehicle project.


\bibliography{IEEEexample}

\end{document}